\documentclass[11pt]{article}
\usepackage{authblk}

\newlength{\defbaselineskip}
\usepackage{graphicx}
\usepackage{amsmath}
\usepackage{amssymb}
\usepackage[caption=false,font=footnotesize]{subfig}
\usepackage{stmaryrd}

\usepackage{algorithmic}
\usepackage{algorithm}

\DeclareMathOperator*{\argmin}{arg\,min}

\begin{document}

\title{
Kernel Ridge Regression Using Importance Sampling with Application to Seismic Response Prediction\footnote{This paper is accepted for publication in IEEE International Conference on Machine Learning and Applications (ICMLA), 2020. 
}}

\author{Farhad Pourkamali-Anaraki
  \\
  \textit{Department of Computer Science, University of Massachusetts Lowell}
  \and
  Mohammad Amin Hariri-Ardebili\\
  \textit{Civil Environmental and Architectural Engineering, University of Colorado Boulder}
  \and 
  Lydia Morawiec\\
   \textit{Department of Electrical \& Computer Engineering, University of Massachusetts Lowell}
}

\date{\vspace{-7ex}}

\maketitle

\begin{abstract}
	Scalable kernel methods, including kernel ridge regression, often rely on low-rank matrix approximations using the Nystr\"om method, which involves selecting landmark points from large data sets. The existing approaches to  selecting landmarks are typically computationally demanding as they require manipulating and performing computations with large matrices in the input or feature space. In this paper, our contribution is twofold. The first contribution is to propose a novel landmark selection method that promotes diversity using an efficient two-step approach. Our landmark selection technique follows a coarse to fine strategy, where the first step computes importance scores with a single pass over the whole data. The second step performs K-means clustering on the constructed coreset to use the obtained centroids as landmarks. Hence, the introduced method provides tunable trade-offs between accuracy and efficiency. Our second contribution is to investigate the performance of several landmark selection techniques using a novel application of kernel methods for predicting structural responses due to earthquake load and material uncertainties. Our experiments exhibit the merits of our proposed landmark selection scheme against baselines.
\end{abstract}

\section{Introduction}\label{sec:into}
Kernel methods provide an effective framework for applying linear models to solve complex nonlinear problems by mapping data into a high-dimensional feature space \cite{hofmann2008kernel}. Well-known examples include support vector machines \cite{cortes1995support,si2017memory}, kernel ridge regression \cite{saunders1998ridge}, and kernel K-means clustering \cite{wang2019scalable}. Kernel-based learning techniques have appeared throughout a wide range of applications, such as analyzing genomic data \cite{huang2018applications} and assessing wind turbine power performance \cite{skrimpas2015employment}.

For a given collection of $n$ data points $\mathcal{X}=\{\mathbf{x}_1,\ldots,\mathbf{x}_n\}$ in $\mathbb{R}^p$, kernel methods compute the inner products in feature space using a kernel function that encodes the pairwise similarities in the input space, for every $\mathbf{x}_i,\mathbf{x}_j\in\mathcal{X}$:
\begin{equation}
[\mathbf{K}]_{ij}:=\kappa(\mathbf{x}_i,\mathbf{x}_j)=\langle\Phi(\mathbf{x}_i),\Phi(\mathbf{x}_j)\rangle=\Phi(\mathbf{x}_i)^T\Phi(\mathbf{x}_j),
\end{equation} 
where $\Phi: \mathbf{x}\mapsto\Phi(\mathbf{x})$ is the kernel-induced feature map. Therefore, the key input to kernel machines is the symmetric positive semidefinite kernel matrix $\mathbf{K}\in\mathbb{R}^{n\times n}$ that measures all the pairwise similarities between the $n$ given data points. A popular choice for the kernel function $\kappa(\cdot,\cdot)$ is the Gaussian radial basis function (RBF) kernel of the following form:
\begin{equation}
\kappa(\mathbf{x}_i,\mathbf{x}_j)=\exp(-d^2(\mathbf{x}_i,\mathbf{x}_j)),\;\;d(\mathbf{x}_i,\mathbf{x}_j):=\|\mathbf{x}_i-\mathbf{x}_j\|_2/\sigma,\label{eq:Gauss}
\end{equation}
with kernel width $\sigma>0$. The advantage of employing such a kernel  as a similarity measure is that it allows constructing algorithms in dot product spaces. However, a critical problem is that kernel methods do not scale favorably with the data size. Forming the entire kernel matrix $\mathbf{K}$ requires $\mathcal{O}(n^2p)$  arithmetic operations to compute scaled Euclidean distances for all pairs of data points. Hence, utilizing kernel methods for large-scale problems is challenging because of the quadratic computational complexity and storage space in $n$. 

Moreover, subsequent processing of the kernel matrix $\mathbf{K}\in\mathbb{R}^{n\times n}$ in kernel-based machine learning methods  tends to be a computationally expensive process. For example, consider the kernel ridge regression problem, which is the main focus of this paper. Let $\{(\mathbf{x}_i,y_i)\}_{i=1}^n$ be $n$ pairs of points in $\mathcal{X}\times\mathcal{Y}$, where $\mathcal{X}$ is the input space and $\mathcal{Y}$ is the response space (we assume $y_i\in\mathbb{R}$). The kernel ridge regression problem boils down to finding the vector $\boldsymbol{\alpha}_{\text{opt}}\in\mathbb{R}^n$ that solves \cite{yasuda2019tight}:
\begin{equation}
\boldsymbol{\alpha}_{\text{opt}}:=\argmin_{\boldsymbol{\alpha}\in\mathbb{R}^n}\|\mathbf{K}\boldsymbol{\alpha}-\mathbf{y}\|_2^2+\lambda \boldsymbol{\alpha}^T\mathbf{K}\boldsymbol{\alpha},
\end{equation}
where $\lambda>0$ is the regularization parameter and $\mathbf{y}=[y_1,\ldots,y_n]^T\in\mathbb{R}^n$ is the target vector. It is known that the above optimization problem has closed form solution:
\begin{equation}
\boldsymbol{\alpha}_{\text{opt}}=(\mathbf{K}+\lambda \mathbf{I}_n)^{-1}\mathbf{y}.\label{eq:alpha_org}
\end{equation}
Using the obtained solution $\boldsymbol{\alpha}_{\text{opt}}$, we estimate the response value for a test data point $\mathbf{x}_{\text{test}}$ as follows:
\begin{equation}
\widehat{y}= \begin{bmatrix} \kappa(\mathbf{x}_{\text{test}},\mathbf{x}_1) & \ldots & \kappa(\mathbf{x}_{\text{test}},\mathbf{x}_n)\end{bmatrix}\boldsymbol{\alpha}.
\end{equation}
Performing kernel ridge regression requires cubic running time concerning the data size $n$ due to the matrix inversion in \eqref{eq:alpha_org}, which is prohibitive in large-scale settings. 

Previous research has focused on exploiting the spectral decay of the kernel matrix $\mathbf{K}$ to reduce the time and space complexities associated with kernel-based machine learning methods. For a target rank parameter $r\leq \text{rank}(\mathbf{K})$, we consider the best rank-$r$ approximation of the kernel matrix $\llbracket  \mathbf{K}\rrbracket_r:=\mathbf{U}_r\boldsymbol{\Lambda}_r\mathbf{U}_r^T$, where $\mathbf{U}_r\in\mathbb{R}^{n\times r}$ and $\boldsymbol{\Lambda}_r\in\mathbb{R}^{r\times r}$ contain the $r$ leading eigenvectors and eigenvalues, respectively. Thus, we can construct the low-rank approximation of the kernel matrix for the target rank $r$ in the following form:
\begin{equation}
\mathbf{K}\approx \llbracket  \mathbf{K}\rrbracket_r=\mathbf{L}\mathbf{L}^T, \;\;\mathbf{L}:=\mathbf{U}_r\boldsymbol{\Lambda}_r^{1/2}\in\mathbb{R}^{n\times r}.\label{eq:low-rank}
\end{equation}
Leveraging the low-rank approximation of the kernel matrix, as shown in \eqref{eq:low-rank}, reduces the memory requirements of kernel methods because the complexity of storing the matrix $\mathbf{L}$ is $\mathcal{O}(nr)$, which is only linear in the number of data points. Also, such a  low-rank approximation leads to noticeable computational savings when analyzing large data sets \cite{zhang2012scaling}. For example, replacing the kernel matrix with its low-rank approximation in \eqref{eq:alpha_org} allows us to find an approximate solution $\widehat{\boldsymbol{\alpha}}$ using the Woodbury inversion lemma \cite{cortes2010impact,pourkamali2018randomized}:
\begin{equation}
\widehat{\boldsymbol{\alpha}}=\big(\mathbf{L}\mathbf{L}^T+\lambda\mathbf{I}_n\big)^{-1}\mathbf{y}=\lambda^{-1}\big(\mathbf{I}_n - \mathbf{L} (\mathbf{L}^T\mathbf{L}+\lambda\mathbf{I}_r)^{-1}\mathbf{L}^T\big)\mathbf{y}.
\end{equation}
Computing this solution takes $\mathcal{O}(nr^2+r^3)$  operations because it requires inverting a much smaller $r\times r$ matrix compared to the solution of  kernel ridge regression given  in \eqref{eq:alpha_org}, resulting in a substantial reduction of time complexity. However, a significant challenge is that computing the exact eigenvalue decomposition of the kernel matrix $\mathbf{K}$ takes at least quadratic time and space concerning the number of data points, which is prohibitively expensive for many applications.

In this paper, we focus on random sampling methods to efficiently generate low-rank approximations in linear time concerning the data size, that we broadly refer to as Nystr\"om approaches \cite{kumar2012sampling}. These methods circumvent the formation of the full kernel matrix by (column) sampling, resulting in reduced time and memory requirements while showing satisfactory empirical performance \cite{gittens2016revisiting}. Given the target rank $r$, the first step of the Nystr\"om method involves selecting $m$ data points, $m\geq r$, from the data set $\mathcal{X}$ according to a probability distribution for generating a set of landmark points $\mathcal{Z}=\{\mathbf{z}_1,\ldots,\mathbf{z}_m\}$. Then, one computes pairwise similarities between the full data set $\mathcal{X}$ and the landmarks $\mathcal{Z}$, as well as  similarities among the elements of $\mathcal{Z}$. Hence, the Nystr\"om method forms two matrices: $\mathbf{C}\in\mathbb{R}^{n\times m}$, i.e., a subset of $m$ columns from $\mathbf{K}$, and  $\mathbf{W}\in\mathbb{R}^{m\times m}$, which is the intersection of the selected columns and their corresponding rows. The Nystr\"om method generates a rank-$m$ approximation of $\mathbf{K}$ in the form of $\mathbf{K}\approx \widehat{\mathbf{K}}=\mathbf{C}\mathbf{W}^\dagger\mathbf{C}^T$ \cite{williams2001using}, where $\mathbf{W}^\dagger$ is the pseudo-inverse of $\mathbf{W}$. The main reason for setting $m$ to exceed the rank parameter $r$ is that increasing the size of the landmark set $\mathcal{Z}$ allows us to extract more information regarding the kernel matrix, resulting in improved rank-$r$ approximations as discussed in recent works \cite{tropp2017fixed,pourkamali2019improved}.

The pivotal aspect of sampling-based low-rank matrix approximations is the landmark selection procedure. The most basic strategy involves sampling the input data points uniformly at random, which does not make use of the available information regarding the data and the similarity measure. Thus, various nonuniform sampling techniques have been proposed to achieve improved trade-offs between accuracy and efficiency, i.e., sampling fewer landmark points to reach a specific accuracy level. A powerful approach builds on selecting \textit{diverse} landmarks that capture data variability in the input space or feature space. However, a significant challenge is the need to process the entire input data set or the kernel matrix. For example, a deterministic technique \cite{zhang2010clustered}, which has shown excellent empirical performance, applies the K-means clustering algorithm to the whole data set $\mathcal{X}$ for finding $m$ cluster centroids as the landmark set. Another recent technique \cite{li2016fast} uses Determinantal Point Processes (DPPs),  discrete probability models that allow the generation of diverse samples. However, this method negates one of the principal benefits of the Nystr\"om method, i.e., avoiding the construction of $\mathbf{K}$. The leverage score sampling technique \cite{alaoui2015fast} also suffers from the same problem as it requires the knowledge of the entire kernel matrix or a high-quality approximation of it \cite{rudi2018fast}, which is not realistic in large-scale data settings.

\begin{figure*}[t]
	\centering
	\includegraphics[width=\linewidth]{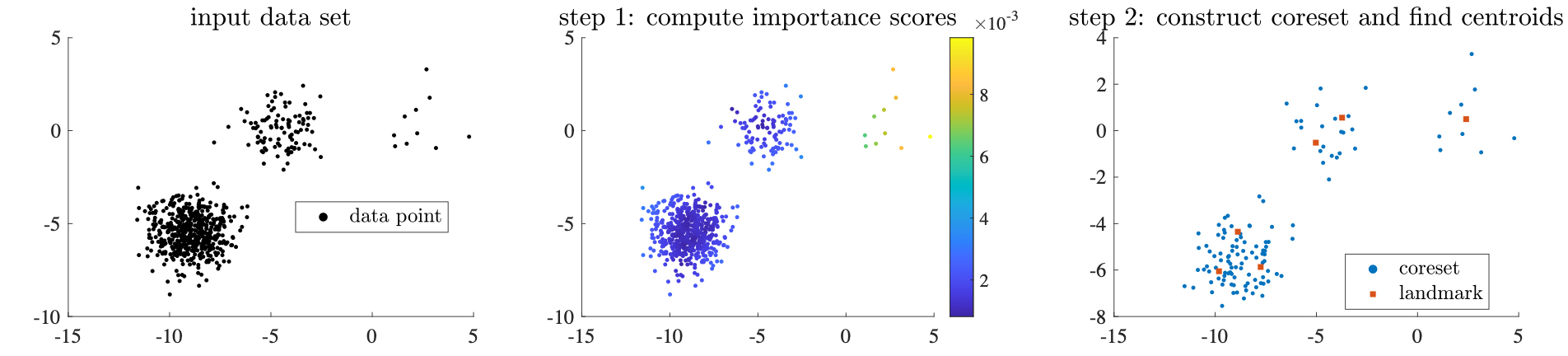}
	\caption{\label{fig:illustration}
		Illustration of the proposed two-step landmark selection technique, which follows a coarse to fine strategy for improving efficiency and accuracy.
	}
\end{figure*}

In this paper, we propose a novel two-step approach to landmark selection for generating Nystr\"om approximations with accuracy-efficiency trade-offs in kernel-based learning, including kernel ridge regression. The proposed landmark selection method is depicted in Fig.~\ref{fig:illustration}. Motivated by recent advances in coreset construction and importance sampling \cite{lucic2017training,feldman2020core}, we design a sampling scheme comprising a \textit{mixture} of uniform and nonuniform components to select a subset of the input data set, often referred to as the coreset. Hence, the first step of our method assigns an importance score to every sample in the input space that tells us how redundant a data point is for the landmark selection purposes. The second step applies K-means clustering to the subset of the original data selected according to the devised distribution, i.e., the coreset, to identify cluster centroids that will form the landmark set in the Nystr\"om approximation. The proposed landmark selection technique requires just a single pass over the whole data set for computing importance scores and relies on the similarity measure introduced by the kernel function. Therefore, our approach significantly reduces the cost of the existing landmark selection technique in  \cite{zhang2010clustered} that performs K-means clustering over the entire data. In contrast to DPP and leverage score sampling, our approach does not need the kernel matrix for selecting landmarks, which is advantageous in large-scale settings. We demonstrate the efficiency and efficacy of our method for selecting diverse landmark points throughout several experiments on synthetic and real data. 

The second main contribution of this paper is that, for the first time, we investigate the performance of the Nystr\"om method and landmark selection techniques for predicting seismic responses of structures with both aleatory and epistemic uncertainty sources. We consider a complex structural analysis that involves uncertainties linked with material properties (e.g., strength and stiffness) and loads (e.g., earthquake magnitude and sea wave height).  Performing nonlinear regression analysis is critical for extracting relationships between input and output parameters using costly continuum level constitutive models, such as finite element simulation codes, across the structural analysis and design community. While the previous research on the Nystr\"om approximation has utilized various benchmark data sets from LIBSVM \cite{CC01a} and UCI Machine Learning Repository, we compare various landmark selection schemes using an application derived from performance-based earthquake engineering. This study further exemplifies the merits of our proposed landmark selection technique based on importance sampling compared to baselines.

The remainder of the paper is outlined as follows. In Section \ref{sec:review}, we present some background, basic tools, and notation. We also review relevant previous landmark selection techniques. In Section \ref{sec:ours}, we describe our landmark selection scheme based on importance sampling and present an intuitive example to exhibit advantages of our approach compared to the prior work. Section \ref{sec:exper} presents numerical experiments on standard benchmark data sets as well as a data set representing a complex structural analysis and modeling in earthquake engineering. We present concluding remarks in Section \ref{sec:conc}. 

\section{Review of the Nystr\"om Method and Existing Landmark Selection Techniques}\label{sec:review}

\subsection{Notation and preliminaries}
We denote column vectors with lower-case bold letters and matrices with upper-case bold letters. We represent the identity matrix of size $n\times n$  by $\mathbf{I}_n$. For a vector $\mathbf{x}$, let $\|\mathbf{x}\|_2$ be the Euclidean norm and $\text{diag}(\mathbf{x})$ represents a diagonal matrix with the elements of $\mathbf{x}$ on the main diagonal. Given a positive semidefinite kernel matrix $\mathbf{K}\in\mathbb{R}^{n\times n}$, we denote the $(i,j)$-th element by $[\mathbf{K}]_{ij}$. The matrix $\mathbf{K}$ admits a factorization, known as the reduced eigenvalue decomposition (EVD), in the form of $\mathbf{K}=\mathbf{U}\boldsymbol{\Lambda}\mathbf{U}^T$, where $\mathbf{U}\in\mathbb{R}^{n\times \rho}$ and $\boldsymbol{\Lambda}=\text{diag}([\lambda_1,\ldots,\lambda_{\rho}])\in\mathbb{R}^{\rho\times \rho}$ contain the eigenvectors and eigenvalues of $\mathbf{K}$, respectively. The parameter $\rho$ denotes the rank of the kernel matrix $\mathbf{K}$ and the eigenvalues are sorted in a non-increasing order $\lambda_1\geq \ldots\geq \lambda_\rho>0$. For any integer $r\leq \rho$, we can form the best rank-$r$ approximation of $\mathbf{K}$ in the form of $\llbracket\mathbf{K}\rrbracket_r=\mathbf{U}_r\boldsymbol{\Lambda}_r\mathbf{U}_r^T$, where $\mathbf{U}_r\in\mathbb{R}^{n\times r}$ represents the first $r$ columns of $\mathbf{U}$, i.e., the $r$ leading eigenvectors, and $\boldsymbol{\Lambda}_r$ contains the $r$ leading eigenvalues on the main diagonal. A partial eigenvalue decomposition of the kernel matrix $\mathbf{K}$ takes $\mathcal{O}(n^2r)$ arithmetic operations. Using this factorization, we can define several standard matrix norms, including the Frobenius norm $\|\mathbf{K}\|_F^2=\sum_{i=1}^{\rho}\lambda_i^2$. The Moore-Penrose pseudo-inverse of $\mathbf{K}$ can be defined using the EVD as $\mathbf{K}^\dagger=\mathbf{U}\boldsymbol{\Lambda}^{-1}\mathbf{U}^T$, where $\boldsymbol{\Lambda}^{-1}=\text{diag}([\lambda_1^{-1},\ldots,\lambda_{\rho}^{-1}])$. When $\mathbf{K}$ is full rank, we get $\mathbf{K}^\dagger=\mathbf{K}^{-1}$. Another factorization that we utilize in this paper is the QR decomposition. For a matrix $\mathbf{C}\in\mathbb{R}^{n\times m}$, with $n\geq m$, we can decompose $\mathbf{C}$ in the form of $\mathbf{C}=\mathbf{Q}\mathbf{R}$, where $\mathbf{Q}\in\mathbb{R}^{n\times m}$ has $m$ orthonormal columns in $\mathbb{R}^n$ and $\mathbf{R}\in\mathbb{R}^{m\times m}$ is an upper triangular matrix. The complexity of the QR decomposition is $\mathcal{O}(nm^2)$. 

The Nystr\"om method is a practical approach for generating approximate low-rank decompositions \textit{without} calculating the whole entries of positive semidefinite kernel matrices and has received significant attention in machine learning \cite{sun2015review}.  The Nystr\"om method first solves a small eigenvalue problem considering the landmark points and then applies an out-of-sample formula to extrapolate the solution to the entire data set. To be formal, the first task is to form $\mathbf{C}\in\mathbb{R}^{n\times m}$ and $\mathbf{W}\in\mathbb{R}^{m\times m}$ using the input data set $\mathcal{X}$, a set of landmark points $\mathcal{Z}$, and the Gaussian RBF kernel. Thus, we get  $[\mathbf{C}]_{ij}=\kappa(\mathbf{x}_i,\mathbf{z}_j)$ and  $[\mathbf{W}]_{ij}=\kappa(\mathbf{z}_i,\mathbf{z}_j)$. Next, the Nystr\"om method utilizes both $\mathbf{C}$ and $\mathbf{W}$ to construct a rank-$m$ approximation of the kernel matrix as $\mathbf{K}\approx \widehat{\mathbf{K}}=\mathbf{C}\mathbf{W}^\dagger\mathbf{C}^T$,  which requires the inversion of $\mathbf{W}$ representing similarities among the landmark points. It is a common practice to set the number of landmarks $m$ to be greater than the target rank parameter $r$ because we expect to acquire more information regarding the structure of the input data as we increase the size of the landmark set. For this reason, the prior work \cite{pourkamali2019improved} presented an efficient technique to restrict the approximation $\widehat{\mathbf{K}}$ to a lower rank-$r$ space. In a nutshell, one computes the QR decomposition of the matrix $\mathbf{C}=\mathbf{Q}\mathbf{R}$, where $\mathbf{Q}\in\mathbb{R}^{n\times m}$ and $\mathbf{R}\in\mathbb{R}^{m\times m}$ as we discussed earlier. This allows us to express the rank-$m$ Nystr\"om approximation in the following form:
\begin{equation}
\widehat{\mathbf{K}}=\mathbf{C}\mathbf{W}^\dagger\mathbf{C}^T=\mathbf{Q}(\mathbf{R}\mathbf{W}^\dagger\mathbf{R}^T)\mathbf{Q}^T=(\mathbf{Q}\mathbf{V})\boldsymbol{\Sigma}(\mathbf{Q}\mathbf{V})^T,
\end{equation}
where we used the EVD of $\mathbf{R}\mathbf{W}^\dagger\mathbf{R}^T=\mathbf{V}\boldsymbol{\Sigma}\mathbf{V}^T$. Note that the columns of $\mathbf{Q}\mathbf{V}\in\mathbb{R}^{n\times m}$ are orthonormal because $\mathbf{V}^T\mathbf{Q}^T\mathbf{Q}\mathbf{V}=\mathbf{V}^T\mathbf{V}=\mathbf{I}_m$. Therefore, the above factorization allows us to find the $r$ leading eigenvectors and eigenvalues of $\widehat{\mathbf{K}}$ as follows: 
\begin{equation}
\widehat{\mathbf{U}}_r^{\text{nys}}:=\mathbf{Q}\mathbf{V}_r,\;\;\widehat{\boldsymbol{\Lambda}}_r^{\text{nys}}:=\boldsymbol{\Sigma}_r. \label{eq:eignys}
\end{equation}
The computational complexity of this technique is linear in the number of input data points $n$ as we perform the EVD on the similarity matrix associated with the landmarks. 

\subsection{Landmark selection techniques}
The most important feature of the Nystr\"om method is the landmark selection process, which influences the approximation error, i.e., $\|\mathbf{K}-\widehat{\mathbf{K}}\|_F$. The selected landmark set also impacts subsequent performance of the approximated kernel-based machine learning algorithms, such as $\|\boldsymbol{\alpha}_{\text{opt}}-\widehat{\boldsymbol{\alpha}}\|_2$ in the case of kernel ridge regression, where $\boldsymbol{\alpha}_{\text{opt}}$ and $\widehat{\boldsymbol{\alpha}}$ represent the solution when utilizing $\mathbf{K}$ and $\widehat{\mathbf{K}}$, respectively. The simplest selection method is uniform sampling without replacement, where each data point is sampled with the same probability, i.e., $p_i=1/n$, for $i=1,\ldots,n$. Despite the simplicity of uniform sampling, a significant downside is that the selected landmark set may not properly represent the underlying structure of the data.  For example, the $m$ landmark points may be redundant or miss critical information regarding some low-density regions within the data set. 

The previous research introduced various sampling strategies to select a diverse or informative subset of the input data. For example, a highly accurate method proposed in  \cite{zhang2010clustered} builds on a simple observation that the centroids obtained by performing the K-means clustering algorithm on the whole data set offer a diverse set of points to summarize the data. To further explain this point, we recall that the K-means clustering objective function is to minimize:
\begin{equation}
\sum_{\mathbf{x}\in\mathcal{X}}\min_{\mathbf{c}\in\mathcal{C}} \|\mathbf{x}-\mathbf{c}\|_2^2,
\end{equation}
where $\mathcal{C}$ is a set of $K$ cluster centroids. Hence, this landmark selection technique sets the number of clusters to $m$ and report the returned set of centroids as the landmark set $\mathcal{Z}$. Since this optimization problem is NP-hard, one should iteratively update assignments and cluster centroids, and typically a few tens of iterations are sufficient \cite{franti2019much}. Therefore, the existing landmark selection method based on K-means clustering requires multiple passes over the whole data, and the computational cost substantially increases for large high-dimensional data. 

Another line of work focuses on a probabilistic framework for selecting diverse landmark points using Determinantal Point Processes (DPPs). In this case, the probability of observing a subset $\mathcal{I}\subset\{1,\ldots,n\}$ is proportional to the determinant of a sub-matrix obtained via the intersection of rows and columns of $\mathbf{K}$ indexed by $\mathcal{I}$ as follows:
\begin{equation}
\text{Pr}(\mathcal{I})=\frac{\text{det}(\mathbf{K}_{\mathcal{I}})}{\text{det}(\mathbf{K}+\mathbf{I}_n)}.
\end{equation}
Sampling from DPPs can be done in polynomial time, but requires constructing the whole kernel matrix $\mathbf{K}$. Thus, sampling landmarks based on DPPs is quite expensive when analyzing large data sets. Under the unrealistic assumption that the kernel matrix is accessible, a recent work \cite{li2016fast} presented an approach based on Gibbs sampling to accelerate the computation of probabilities. 

In this paper, we compare our proposed landmark selection approach with the two prior techniques described in this section. The main reason is that these two methods \textit{directly} aim to select diverse landmarks similar to ours. However, there is another probabilistic technique known as leverage score sampling \cite{alaoui2015fast}. This method uses what are known as the $\lambda$-ridge leverage scores for the kernel ridge regression problem. The idea is to select $m$ landmark points with probabilities proportional to the diagonal entries of $\mathbf{K}(\mathbf{K}+\lambda \mathbf{I}_n)^{-1}$. Like DPP sampling, the exact computation of this quantity is as expensive as solving the original kernel ridge regression problem. Thus, several methods have considered different strategies to reduce the cost of finding leverage scores; see \cite{rudi2018fast} for a review.  Based on the reported results in \cite{li2016fast}, DPP sampling provides better accuracy-efficiency trade-offs compared to the leverage score sampling. 

\section{The Proposed Landmark Selection Method}\label{sec:ours}
In this section, we present a two-step approach that can be viewed as a \textit{probabilistic} modification of the prior landmark selection method based on K-means clustering. To this end, we follow a coarse to fine strategy and build on recent advances in importance sampling. The key idea is that we only need approximate solutions to K-means clustering for generating landmark points, allowing us to utilize a subset of the original data instead of processing the whole input data.

The first task is to devise a sampling mechanism to identify a small subset of the data, known as the coreset, that allows obtaining cluster centroids that are competitive with performing K-means clustering on the whole data set. A significant challenge is to select representative points from all clusters present in the data and remove redundant samples. For example, it is known that uniform sampling is biased towards dense regions of the input data, which means that the sampling process may miss critical information regarding the underlying structure of the input data. Thus, we often observe that nonuniform landmark selection techniques outperform uniform sampling for small landmark sets. The second step of our approach involves performing K-means clustering on the constructed coreset to produce the landmark set for generating the Nystr\"om approximation. The proposed method is illustrated in Fig.~\ref{fig:illustration}, and we explain the procedure to compute importance sampling scores in the following. 

\begin{figure*}[t]
	\centering
	\subfloat[][\label{fig:exp1-uni}Uniform Sampling]{
		\includegraphics[width=0.5\linewidth]{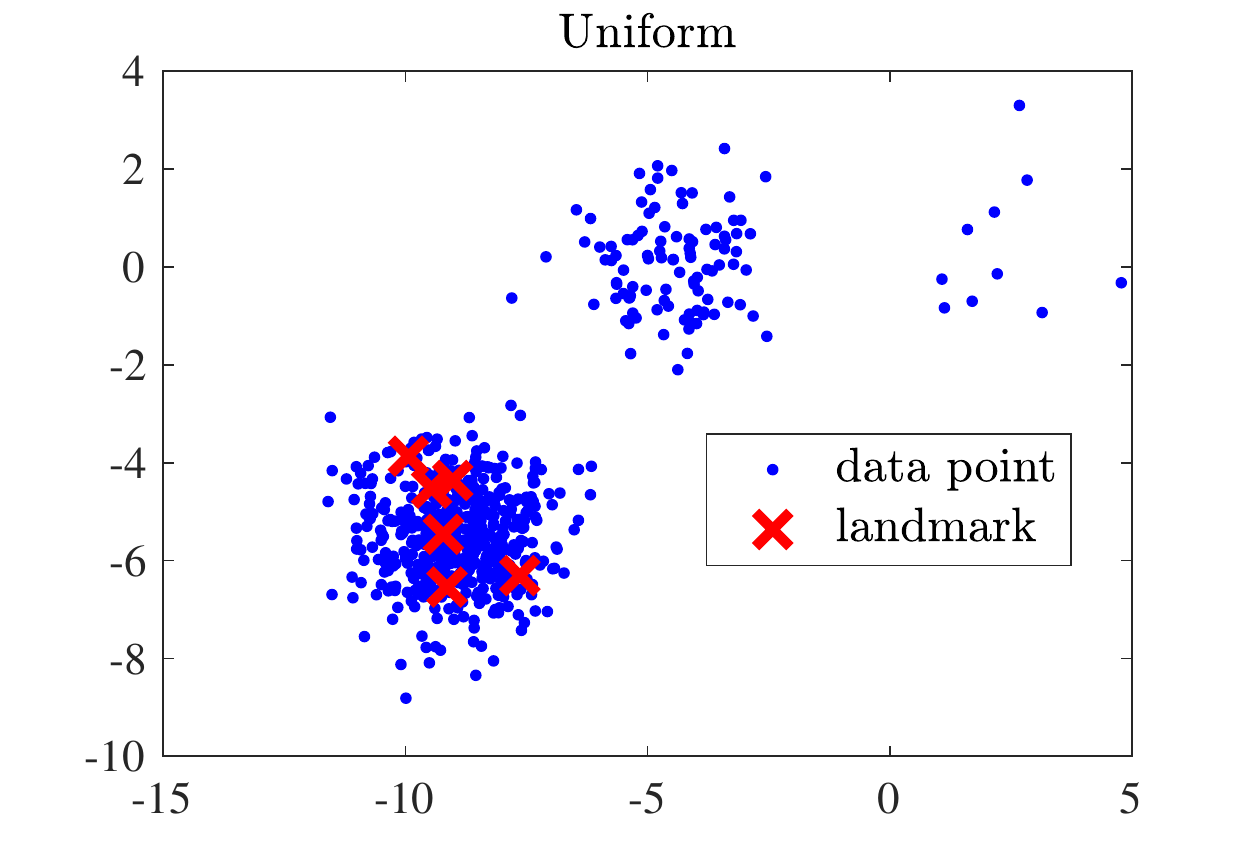}
	}
	\subfloat[][\label{fig:exp1-dpp}DPP Sampling]{
		\includegraphics[width=0.5\linewidth]{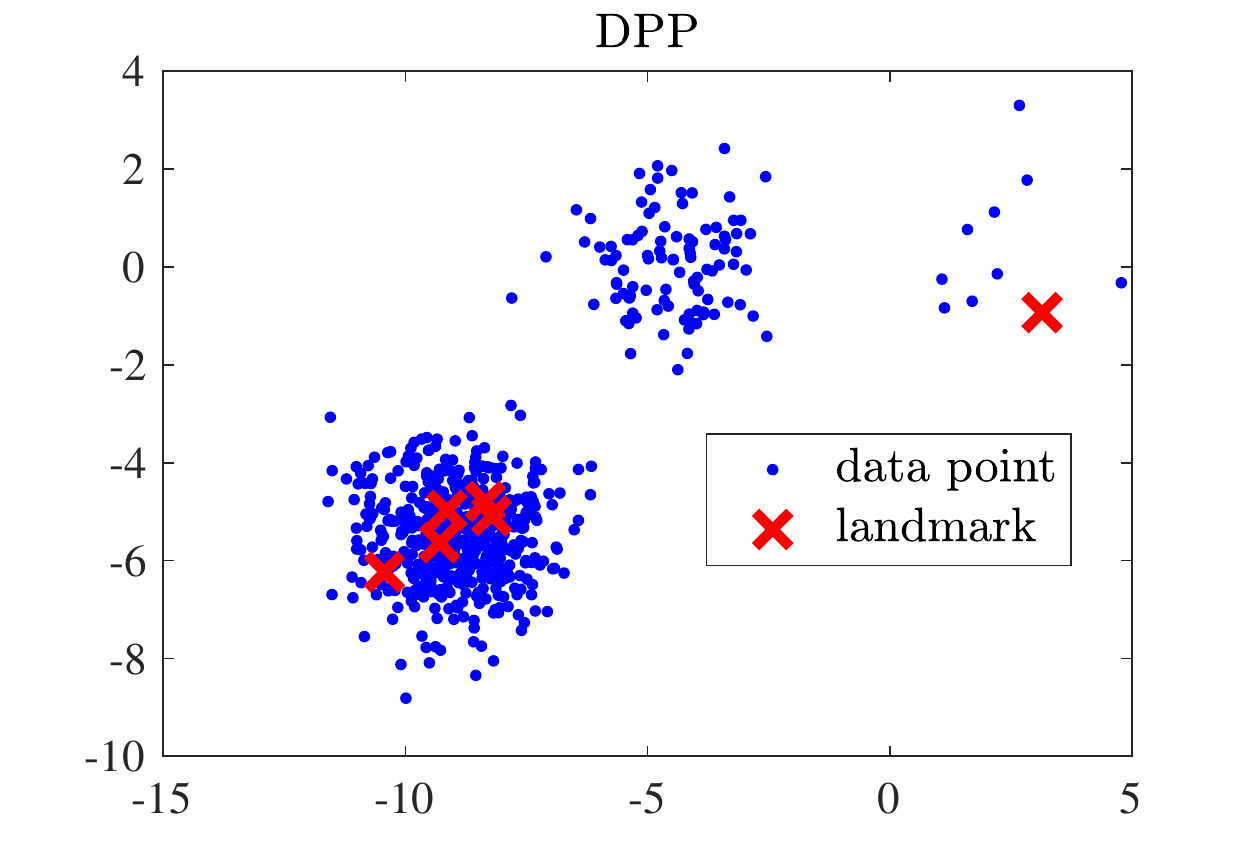}
	} 
	
	\subfloat[][\label{fig:exp1-kmeans}K-means]{
		\includegraphics[width=0.5\linewidth]{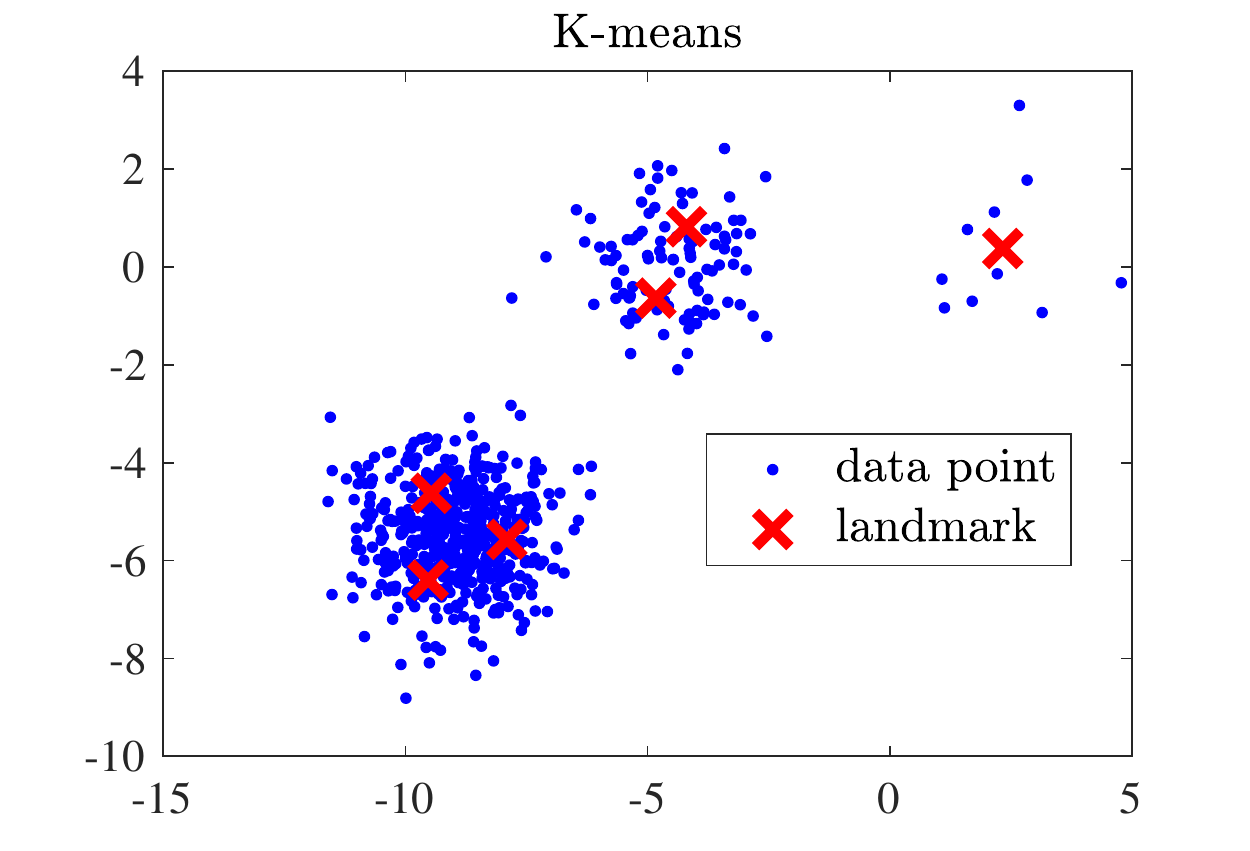}
	}
	\subfloat[][\label{fig:exp1-ours}Importance Sampling (Ours)]{
		\includegraphics[width=0.5\linewidth]{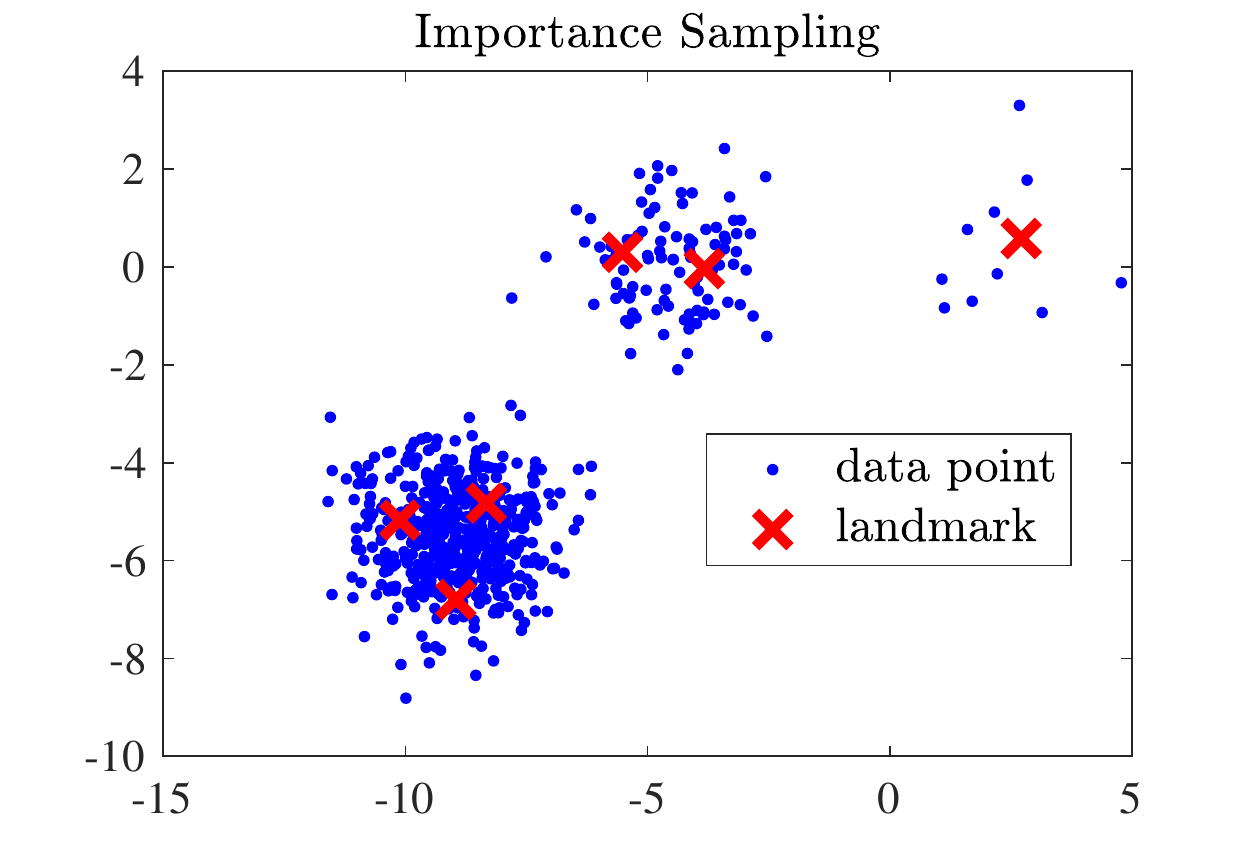}
	} 
	\caption{\label{fig:exper1}
		Exploring the effectiveness of various landmark selection strategies on the synthetic imbalanced data with $m=6$ landmarks.
	}
\end{figure*}

The proposed method starts with a small subset of the input data $\mathcal{S}_0\subset\mathcal{X}$ of size $|\mathcal{S}_0|=n_0$. For example, one can pick the elements of $\mathcal{S}_0$  by sampling uniformly at random without replacement. Next, we compute the distance between each data point $\mathbf{x}\in\mathcal{X}$ and the set $\mathcal{S}_0$ using the distance metric induced by the Gaussian RBF function in \eqref{eq:Gauss}:  
\begin{equation}
d(\mathbf{x},\mathcal{S}_0):=\min_{\mathbf{x}'\in\mathcal{S}_0}d(\mathbf{x},\mathbf{x}'),\;\;d(\mathbf{x},\mathbf{x}')=\|\mathbf{x}-\mathbf{x}'\|_2/\sigma,
\end{equation}
where $\sigma>0$ is the kernel width parameter. Using the computed distances, we propose the following importance sampling score for each data point $\mathbf{x}\in\mathcal{X}$:
\begin{equation}
p(\mathbf{x}):=\underbrace{\frac{1}{2n}}_{\text{part 1: uniform}}+\underbrace{\frac{1}{2}\frac{d(\mathbf{x},\mathcal{S}_0)}{\sum_{\mathbf{x}'\in\mathcal{X}}d(\mathbf{x}',\mathcal{S}_0)}}_{\text{part 2: nonuniform}},\label{eq:samp}
\end{equation}
which is a mixture of two components: (1) uniform sampling that ensures choosing every data point with a nonzero probability and (2) nonuniform sampling that places more weight on selecting points that are distant from the initial set $\mathcal{S}_0$. It is straightforward to show that $p(\mathbf{x})$ is a valid probability distribution since $\sum_{\mathbf{x}\in\mathcal{X}}p(\mathbf{x})=1$ and $p(\mathbf{x})\geq 0$.

The last step of our landmark selection scheme samples $n_1<n$ points from $\mathcal{X}$ according to the importance sampling scores  $p(\mathbf{x})$ given in \eqref{eq:samp} to construct the coreset $\mathcal{S}_1$. We then perform K-means clustering on $\mathcal{S}_1$ and set the number of clusters to $m$, which will result in $m$ cluster centroids. 

The main advantage of the proposed method compared to the prior work is that we no longer need to perform K-means clustering on the whole data or compute the entire kernel matrix. Our landmark selection technique requires just one pass over the data to calculate all distances, and then we sample a fraction of data points using the computed importance sampling scores. Hence, the cost of performing K-means clustering on the reduced data is sub-linear concerning the number of input data points, a significant computational gain compared to the related work for large data sets.

To further explain the merits of our landmark selection method, we consider a synthetic data set with a disproportionate number of points in each region (shown in Fig.~\ref{fig:illustration}). This data set contains $n=610$ samples in $\mathbb{R}^2$, and we choose the kernel parameter $\sigma$ based on the average distance between data points and the sample mean, which is a popular strategy for choosing $\sigma$ in the previous research, e.g., \cite{zhang2010clustered,wang2019scalable}. We also set $n_0=10$, $n_1=\lfloor0.2n\rfloor$, and the target rank $r=2$. In Fig.~\ref{fig:exper1}, we visualize $m=6$ produced landmark points using various landmark selection methods. As expected, uniform sampling is biased towards dense regions of $\mathcal{X}$ and fails to represent the underlying distribution of the input data. On the other hand, K-means clustering and our approach provide a reasonable landmark set, covering the entire data set. Unfortunately, we observe that DPP, which is designed to select diverse landmarks, misses critical information regarding one of the clusters present in the data.

\begin{figure}[ht!]
	\centering
	\includegraphics[width=0.55\linewidth]{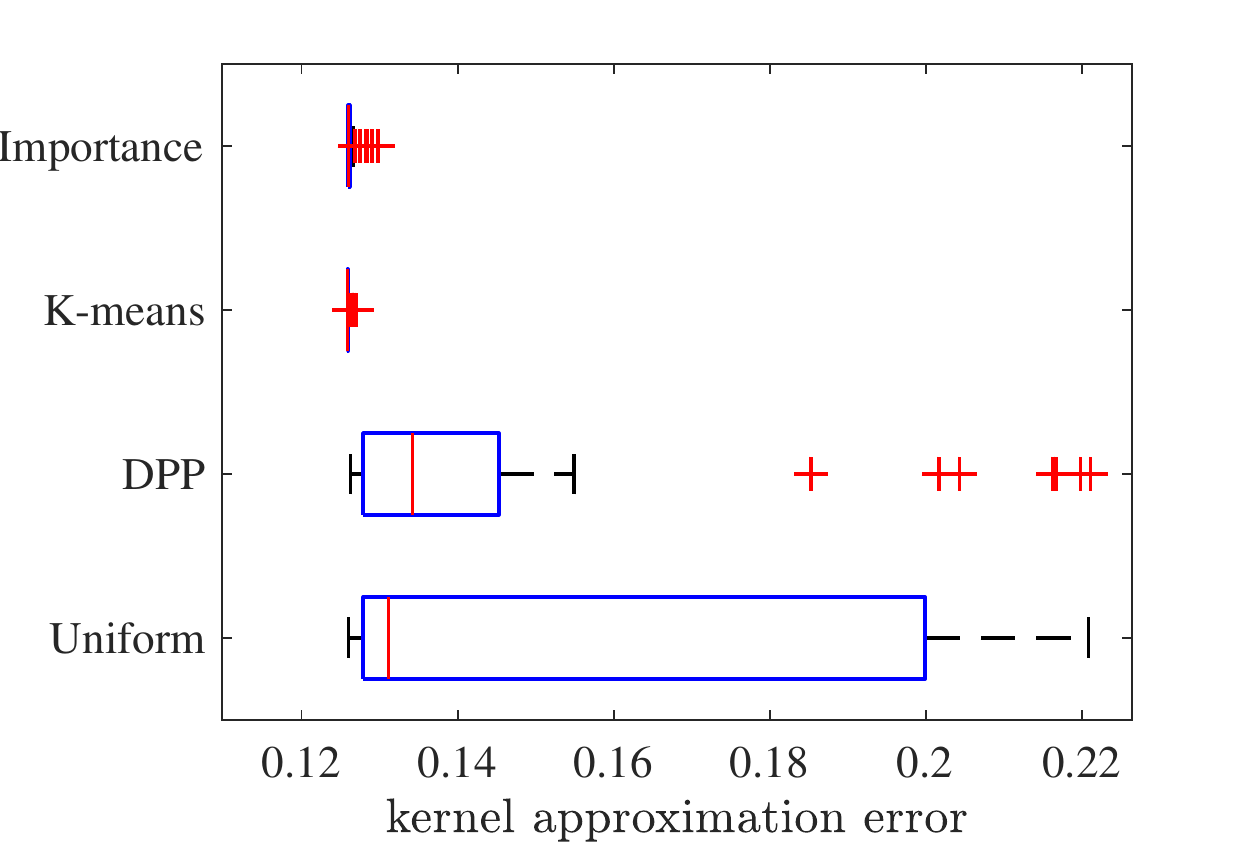}
	\caption{\label{fig:exper2}
		Reporting the normalized kernel approximation error on the synthetic data over $50$ trials for the target rank $r=2$ and $m=6$ landmarks.
	}
\end{figure}

 Moreover, we evaluate the performance of these techniques using the normalized kernel approximation error, which is defined based on the estimated eigenvalues and eigenvectors of the Nystr\"om approximation explained in \eqref{eq:eignys}:
 \begin{equation}
 \text{approx err}= \frac{\|\mathbf{K} - \widehat{\mathbf{U}}_r^{\text{nys}}\widehat{\boldsymbol{\Lambda}}_r^{\text{nys}}(\widehat{\mathbf{U}}_r^{\text{nys}})^T\|_F}{\|\mathbf{K}\|_F}. \label{eq:approxerr}
 \end{equation}
  Fig.~\ref{fig:exper2} reports the approximation error over $50$ independent trials, which further demonstrates that our method generates Nystr\"om approximations using the proposed importance sampling technique with negligible accuracy and performance loss. As a baseline, the normalized approximation error obtained by computing the exact EVD of $\mathbf{K}$ is $0.126$ in this example.
  
\section{Experimental Results}\label{sec:exper}
In our experiments, we evaluate the accuracy and running time of the proposed landmark selection method, referred to as Importance Sampling, through  kernel approximation and kernel ridge regression tasks. We consider two benchmark data sets from LIBSVM \cite{CC01a}, namely satimage and cadata.  We also study a novel application of kernel ridge regression in the field of earthquake and structural engineering. To our knowledge, we explore,  for the first time, the effectiveness of landmark selection techniques in the context of the Nystr\"om approximation in engineering domains. Therefore, our results provide a road map for the future work to scale up kernel-based learning methods for large-scale analysis in engineering.

In our implementation, we use standard internal functions in Matlab to perform matrix factorization, such as eigenvalue decomposition (EVD) and QR decomposition. We also employ the Matlab's built-in K-means clustering algorithm. For the prior work on selecting landmarks based on K-means and our method, which involves clustering coresets, we set the number of K-means iterations to $20$ as we did not observe any noticeable improvement by increasing this parameter. We also fix the size of the initial set $n_0=|\mathcal{S}_0|=10$ since adjusting this parameter does not affect much the performance of our landmark selection technique. We will present an experiment demonstrating the influence of $|\mathcal{S}_1|$, i.e., size of coresets, on the approximation quality of kernel matrices. We use the Matlab code provided by the authors to implement DPP sampling \cite{li2016fast}, which uses Gibbs sampling for improving efficiency. 

Throughout this section, we use varying values of the target rank $r$ and the number of landmarks $m$ to perform an extensive comparison of landmark selection strategies.  In all experiments, we choose the kernel width parameter $\sigma$ based on the average distance between data points and the sample mean. We did not use a fine-tuning process for the kernel parameter $\sigma$ to have a fair comparison with the previous work. To reduce the statistical variability, experiments involving randomness in the sampling process are repeated $50$ times, and the average results with standard deviations are reported.

\subsection{Kernel matrix approximation}
\begin{figure*}[t]
	\centering
	\subfloat[][\label{fig:exper_app1}Kernel approximation error]{
		\includegraphics[width=0.5\linewidth]{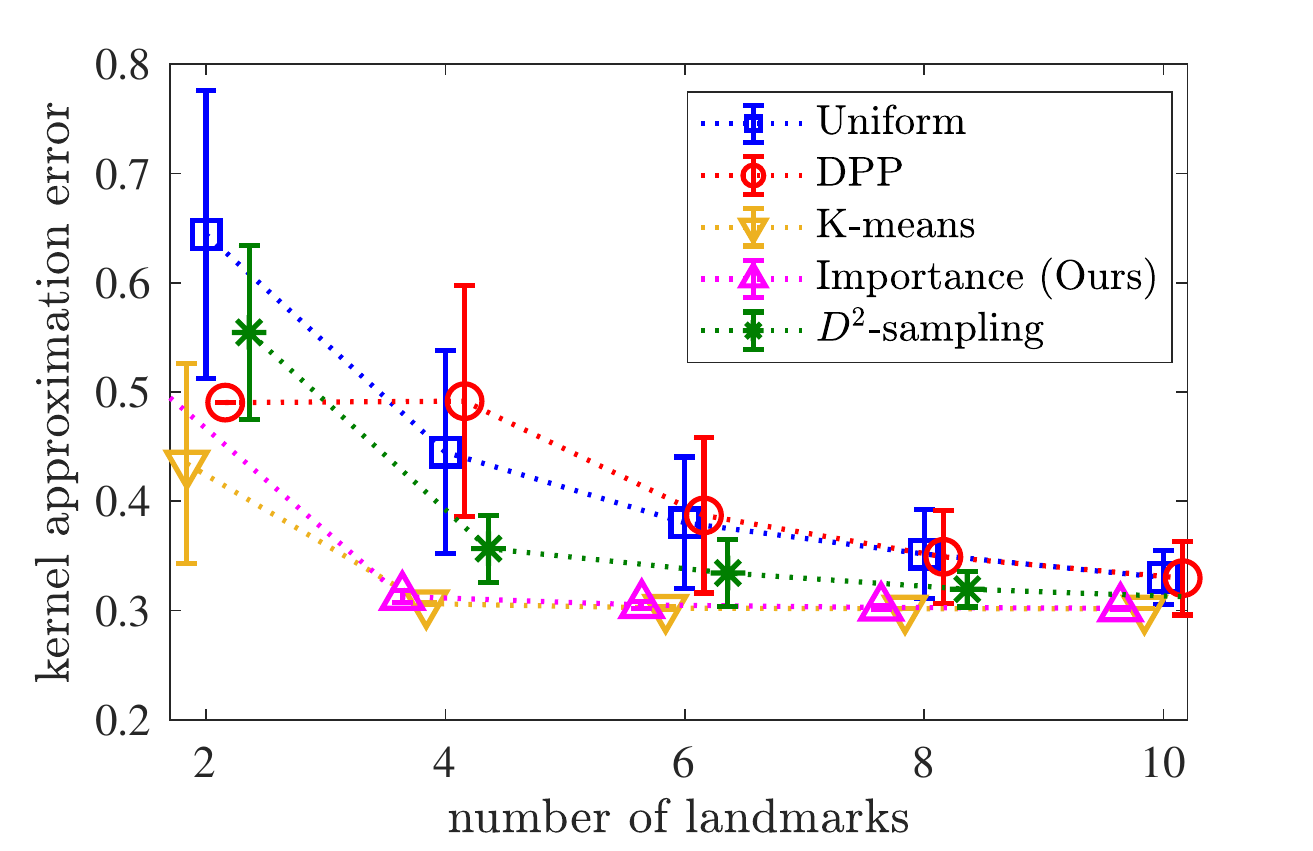}
	}
	\subfloat[][\label{fig:exper_app2}Time complexity]{
		\includegraphics[width=0.5\linewidth]{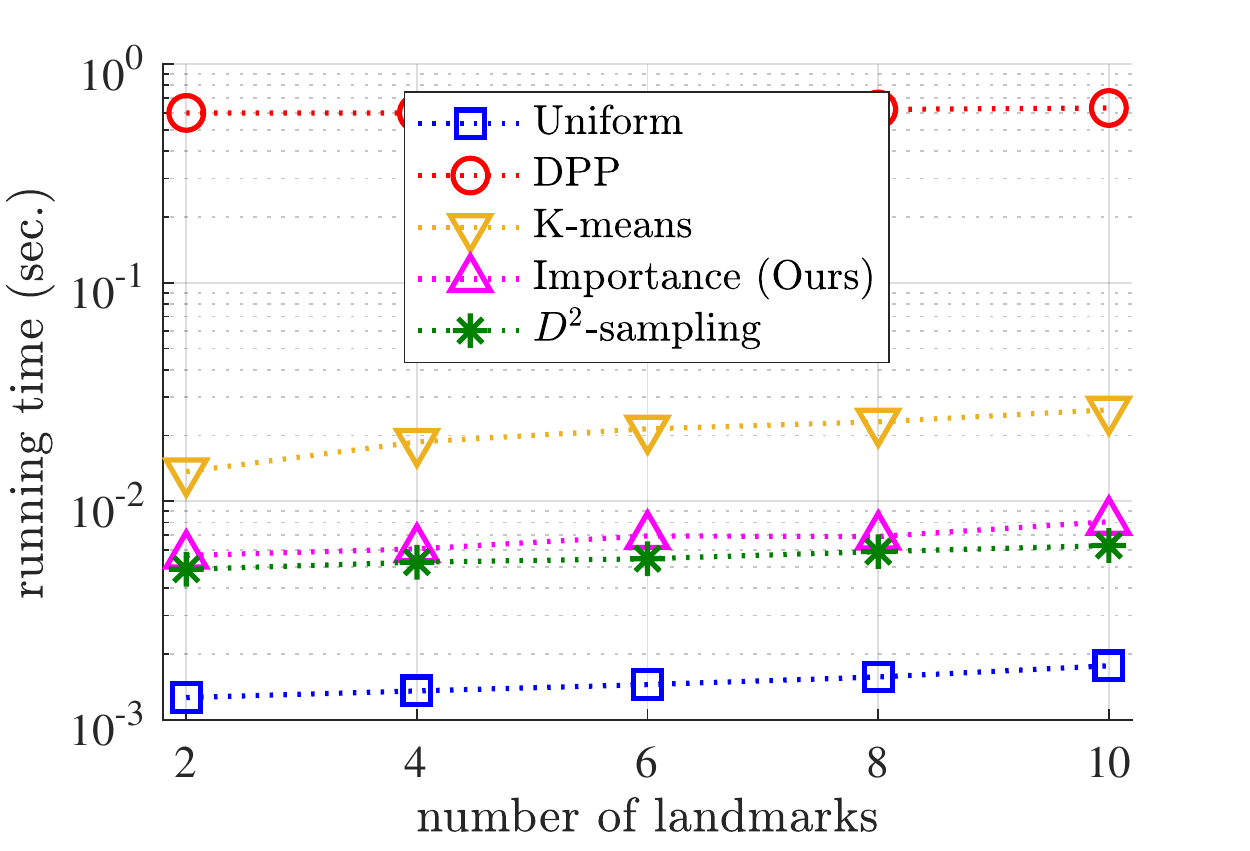}
	}
	
	\caption{\label{fig:exper_app}
		Kernel matrix approximation error and runtime on the satimage data set.
	}
\end{figure*}

The first experiment examines the kernel matrix approximation error, defined in \eqref{eq:approxerr}, on the satimage data set for fixed rank $r=2$ and varying numbers of landmark points $m=r,\ldots,5r$. This data set contains $n=10,\!870$ samples in $\mathbb{R}^p$, where $p=36$. When forming the whole kernel matrix and computing the exact EVD, the approximation error is $0.302$. Fig.~\ref{fig:exper_app1} presents the mean and standard deviation of the approximation error. These results show that, for $m\geq 4$, the accuracies of the previous work based on K-means and our importance sampling method reach the accuracy of the best rank-$2$ approximation obtained by the exact EVD. We also see that DPP sampling does not consistently outperform uniform sampling. To demonstrate the effectiveness of the introduced sampling distribution in \eqref{eq:samp}, we also consider a widely-used sensitivity score known as $D^2$-sampling \cite{kmeansplus,bachem2018one}, which defines the following distribution over the data:
\begin{equation}
q(\mathbf{x}):=\frac{d^2(\mathbf{x},\mathcal{S}_0)}{\sum_{\mathbf{x}'\in\mathcal{X}}d^2(\mathbf{x}',\mathcal{S}_0)},\;\;\text{for }\mathbf{x}\in\mathcal{X}.
\end{equation}
Compared to our approach, this sampling distribution only consists of a nonuniform term, which depends on the squared distances between data points and $\mathcal{S}_0$. In Fig.~\ref{fig:exper_app1}, we observe that utilizing our proposed importance sampling provides improved Nystr\"om approximations. In terms of computational complexity, we present timing results in Fig.~\ref{fig:exper_app2}, revealing that the uniform sampling technique is faster than the other methods. On the other hand, DPP sampling suffers from high computational cost because of forming the entire kernel matrix $\mathbf{K}$. We also see that our approach reduces the time complexity of the previous work on selecting landmarks based on K-means clustering. We only need to perform K-means on coresets, and computing importance scores is straightforward. Overall, our proposed method provides the best trade-off between accuracy and efficiency in this experiment. 

\subsection{Kernel ridge regression}
This experiment investigates the impact of utilizing a rank-$r$ approximation of $\mathbf{K}$ on the performance of kernel ridge regression that we discussed in Section \ref{sec:into}. For this task, we use a data set with continuous target values from LIBSVM. The data set, named cadata, contains $n=20,\!640$ samples in $\mathbb{R}^8$ and we use $70\%$ of the whole data set for training kernel ridge regression and the remainder for evaluating the performance, based on the coefficient of determination:
\begin{equation}
1-\frac{\sum_{i=1}^{n} (y_i-\widehat{y}_i)^2}{\sum_{i=1}^n (y_i - \bar{y})^2},\;\;\text{and }\bar{y}:=\frac{1}{n}\sum_{i=1}^{n}y_i,
\end{equation}
where $\widehat{y}_i$, $i=1,\ldots,n$, are predicted values obtained by a trained regression model. Best possible score is $1$ and it can be negative. Thus, higher values of the coefficient of determination indicate more accurate regression models.

\begin{figure*}[t]
	\centering
	\subfloat[][\label{fig:exper_reg1}Coefficient of determination]{
		\includegraphics[width=0.5\linewidth]{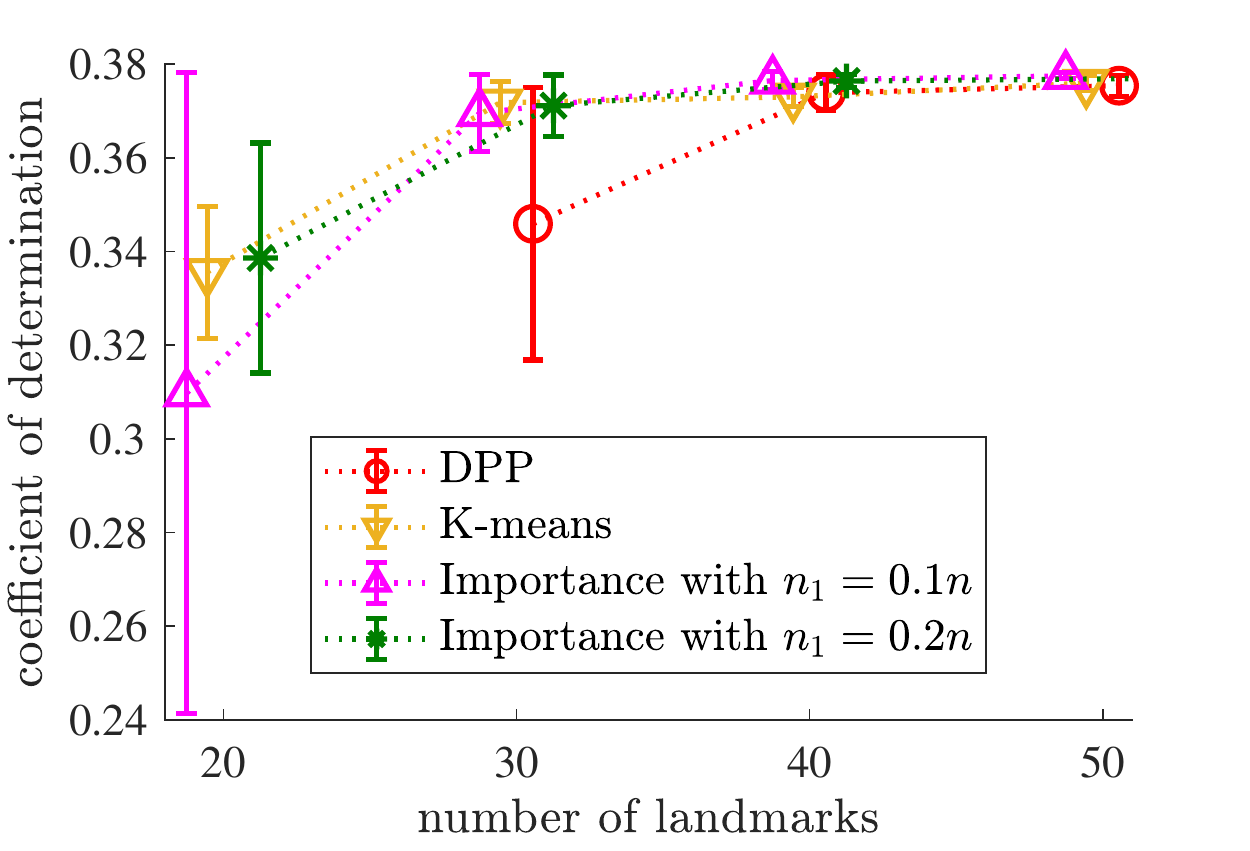}
	}
	\subfloat[][\label{fig:exper_reg2}Time complexity]{
		\includegraphics[width=0.5\linewidth]{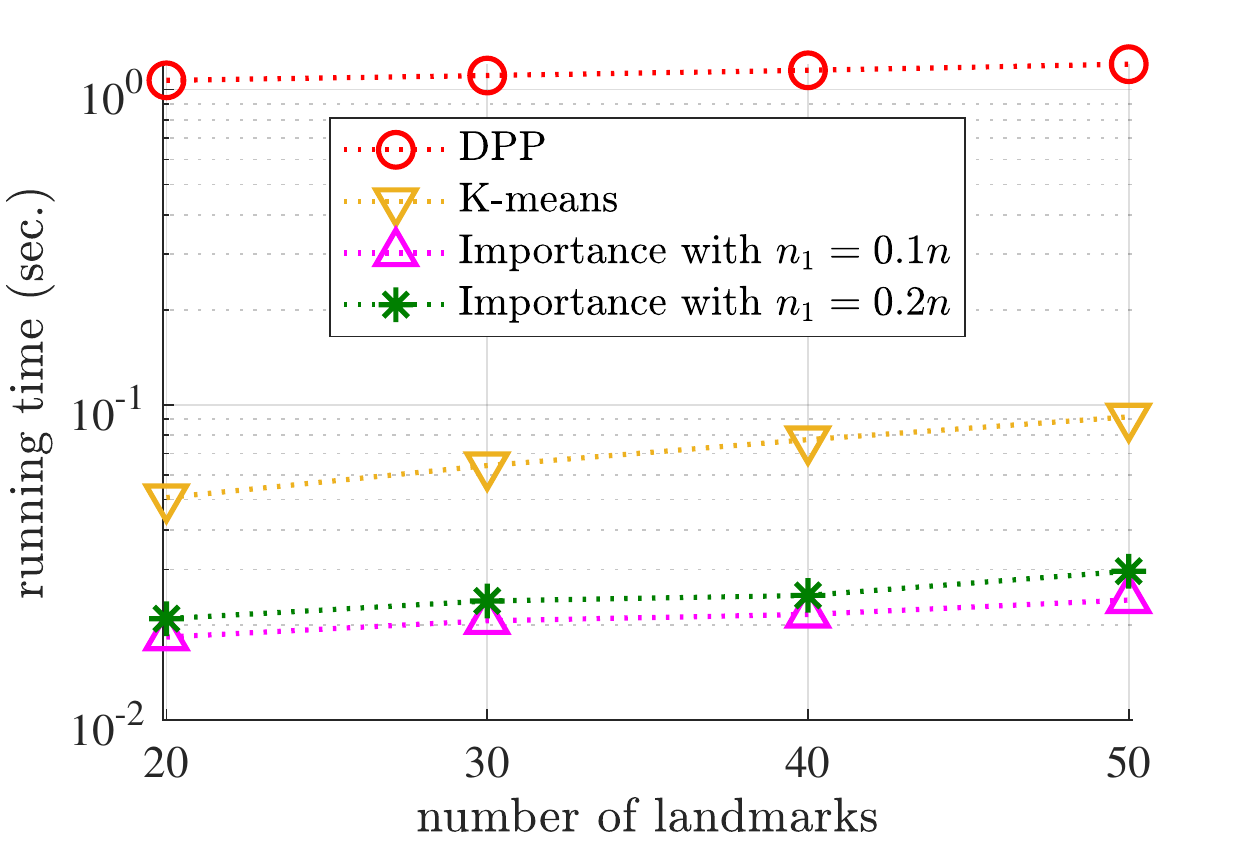}
	}
	
	\caption{\label{fig:exper_reg}
		Accuracy and running time of kernel ridge regression on cadata for varying landmark points and two coreset sizes. 
	}
\end{figure*}

In Fig.~\ref{fig:exper_reg1}, we report the mean and standard deviation of the coefficient of determination for fixed $r=20$, $\lambda=1$, and varying values of landmark points ranging from $20$ to $50$. We omit the results for uniform sampling (and DPP when $m=20$) as the corresponding values of the coefficient of determination are negative, i.e., they are worse than a model that returns a fixed value for any input data point. We see that our method's performance using coresets of size $n_1=\lfloor 0.2n\rfloor$ is on par with the K-means approach. Also, decreasing the size of coresets $\mathcal{S}_1$, such as $n_1=\lfloor 0.1n\rfloor$, impacts our approach's accuracy just for $m=20$. Furthermore, we see that the accuracy of DPP is significantly improved for larger values of $m$, e.g., $m=40$. Similar to the previous experiment, Fig.~\ref{fig:exper_reg2} reveals that our method is more efficient than both DPP and K-means clustering. Also, increasing the size of coresets does not notably increase the time complexity of our method because of the fixed cost of forming $\mathbf{C}$ and $\mathbf{W}$.

\subsection{Application: seismic response prediction}

Uncertainty quantification is a critical task in risk-based safety management of engineering structures. The performance-based earthquake engineering framework proposed by the Pacific Earthquake Engineering Research Center (PEER) is the widely-used approach for safety assessment of different structural systems. This framework accounts for both aleatory (resulted from ground motion record-to-record variability) and epistemic (mainly from material and modeling randomness) uncertainties. Multiple studies reported the hybrid impact of these uncertainty sources on the overall dispersion of results \cite{dolsek2009incremental}. Since the conventional combination of various uncertainty sources is computationally expensive, an active line of research aims to develop data-driven methods for reducing the overall number of simulations/experiments without compromising accuracy \cite{hariri2019efficient}.

\begin{figure*}[t]
	\centering
	\subfloat[][\label{fig:exper_earth0}Schematic diagram of our model]{
		\includegraphics[width=0.33\linewidth]{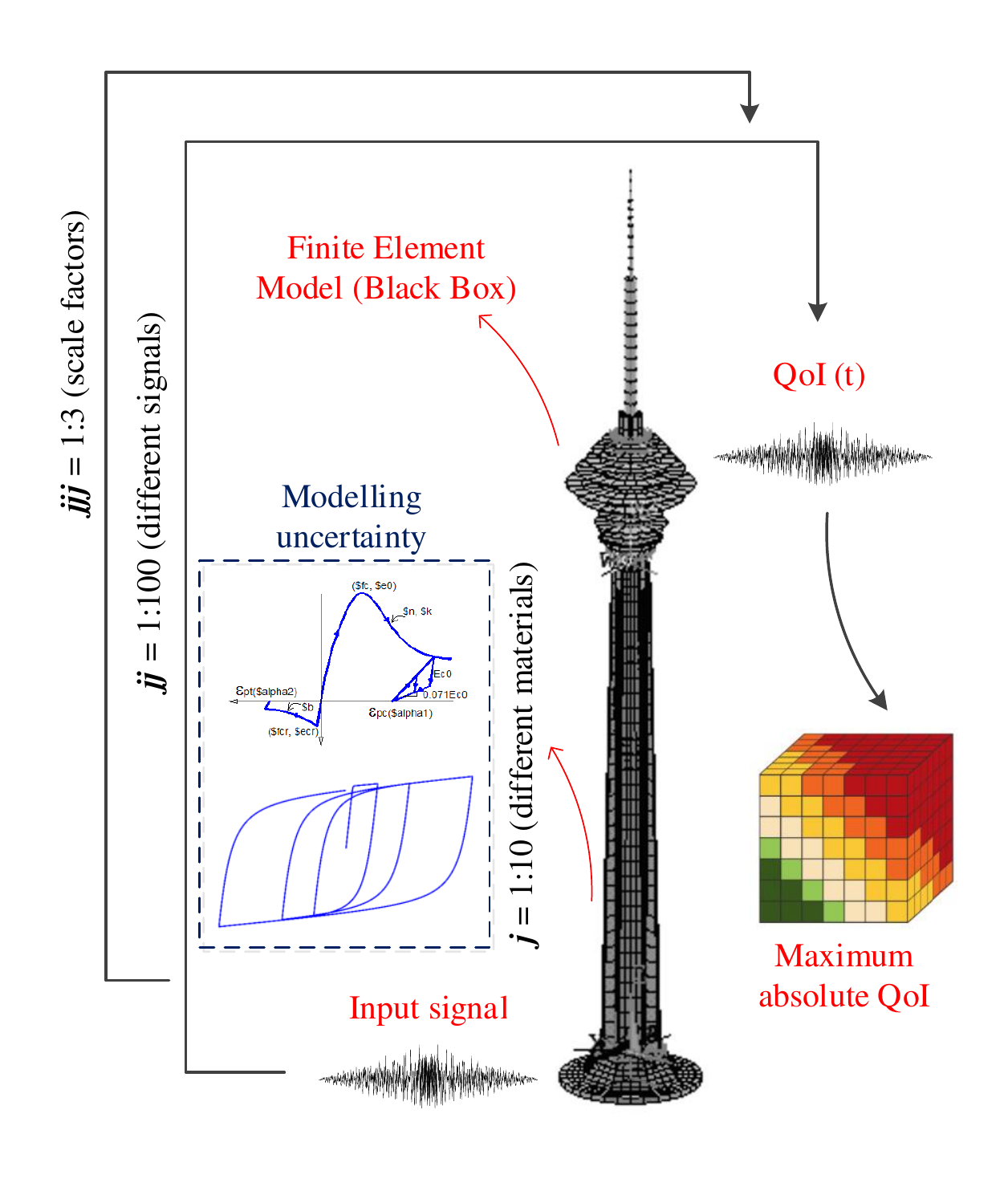}
	}
	\subfloat[][\label{fig:exper_earth1}Top displacement]{
		\includegraphics[width=0.55\linewidth]{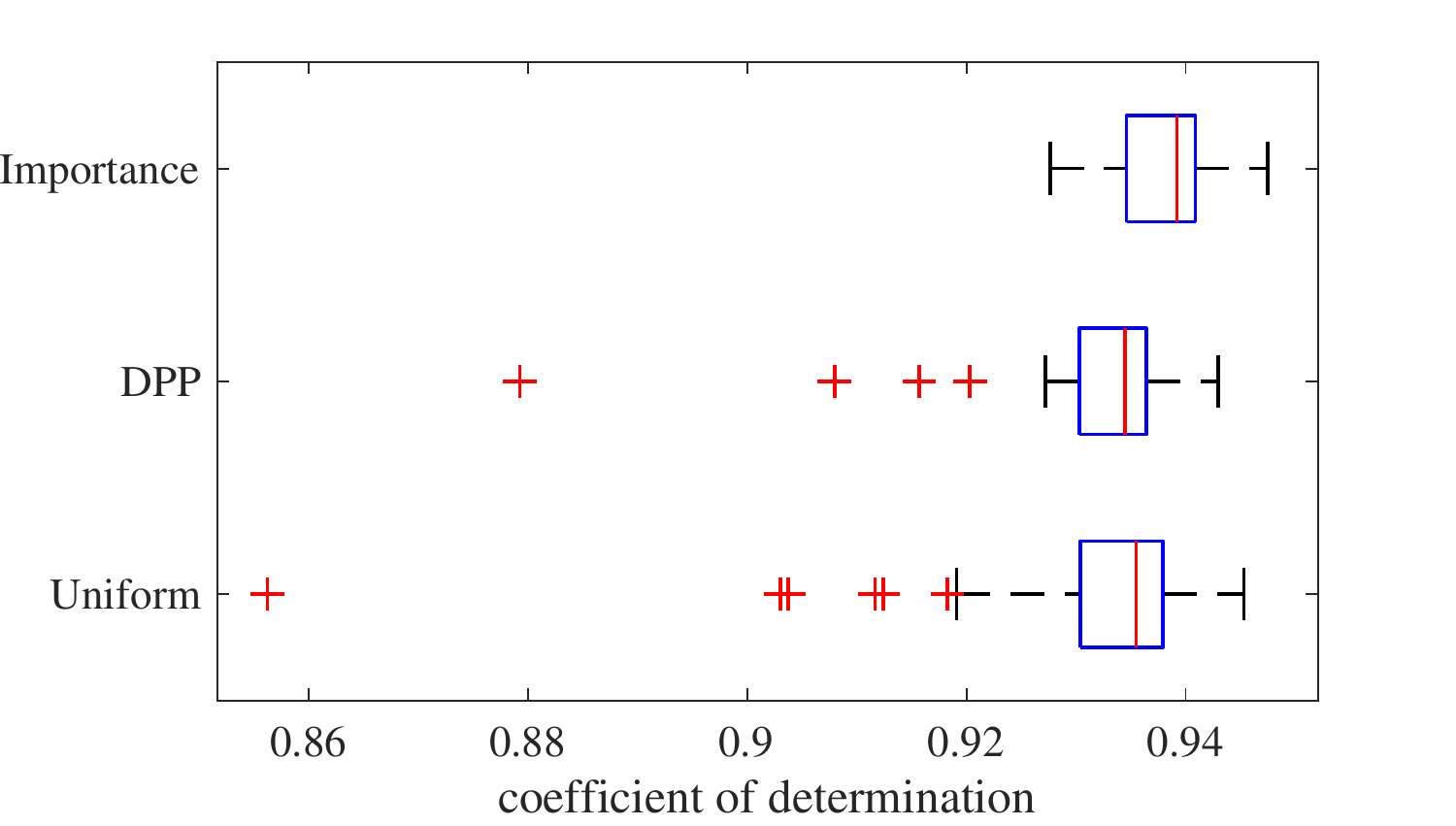}
	}

	\subfloat[][\label{fig:exper_earth2}Base shear]{
		\includegraphics[width=0.55\linewidth]{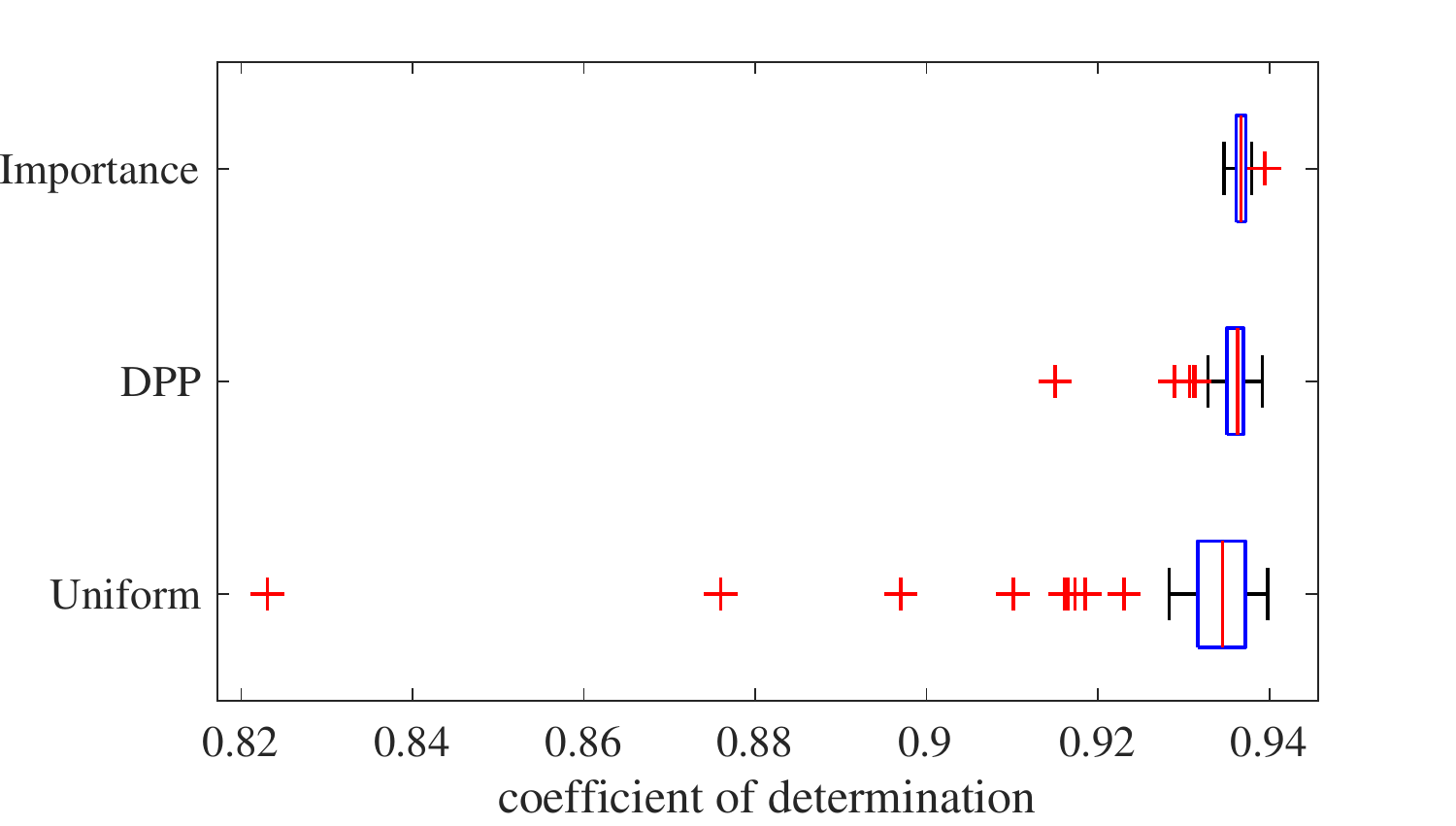}
	}
 
	\caption{\label{fig:exper_earth}
	Comparing accuracy of various landmark selection techniques on a data set derived from the finite element analysis of a tower. 
	}
\end{figure*}

In this paper, we consider a high-rise telecommunication tower as a case study \cite{HaririSamaniMirtaheri2014}. The height is over $400$ meters, made of reinforced concrete. The concrete shaft is the main load-carrying structure of the tower that transfers the lateral and gravitational loads to the foundation. We consider several modeling aspects, including material nonlinearities (i.e., cracking, crushing, and damage), and geometric nonlinearities. We present a schematic 3D finite element model in Fig.~\ref{fig:exper_earth0}. We develop the simplified 2D model of the tower, including the head structure, shaft, and transition. A total of $10$ random models are generated using Latin Hypercube Sampling (LHS), to consider the variability in 18 material/modeling parameters (concrete, steel, and system level). Moreover, $100$ ground motions are used to account for aleatory uncertainty. For each ground motion, we extract $31$ intensity measure parameters \cite{HaririSaoumaCollapseFragility}, including all peak values (e.g., PGA), intensity-, frequency-, and duration-dependent parameters. Overall, we create a data set containing $n=3,\!000$ simulations with $p=49$ attributes. The output space for the regression analysis represents two structural responses: top displacement and base shear. 

In this experiment, we compare the performance of our approach against baselines for predicting the two structural responses. We use $n=2,\!500$ input-output pairs for training and the remaining $500$ examples for evaluating the accuracy of trained models. We set the rank parameter $r=10$, the number of landmarks $m=20$, the regularization parameter $\lambda=1$, and the size of coreset $n_1=\lfloor0.1n\rfloor$. We show the values of the coefficient of determination over $50$ independent trials in Fig.~\ref{fig:exper_earth1} and Fig.~\ref{fig:exper_earth2} for top displacement and base shear, respectively. The previous landmark selection method based on K-means clustering resulted in negative values, so we omitted to improve the readability of these results. We tried to enhance this method's accuracy by increasing the number of iterations, which was not successful. The reported results show that the accuracy of uniform sampling varies significantly across different trials, which is problematic in practice. Moreover, we see that our proposed landmark selection method consistently results in accurate regression models and outperforms DPP sampling for predicting both quantities of interest. 

\section{Conclusion}\label{sec:conc}
This work presented a novel landmark selection approach based on constructing coresets using an importance sampling method. Comprehensive experiments on  benchmark data sets and a new application in structural engineering have shown the advantages of our approach in terms of accuracy and efficiency. The proposed method is valuable for accelerating kernel methods to facilitate the development of scientific machine learning techniques.

\bibliographystyle{ieeetr}
\bibliography{sample}

\end{document}